\documentclass[10pt,twocolumn,letterpaper]{article}

\usepackage{iccv}
\usepackage{times}
\usepackage{epsfig}
\usepackage{graphicx}
\usepackage{amsmath}
\usepackage{amssymb}
\usepackage{booktabs}
\usepackage{multirow}
\usepackage{threeparttable}
\usepackage{float}
\usepackage{url}
\usepackage[normalem]{ulem}
\usepackage[accsupp]{axessibility}  % Improves PDF readability for those with disabilities.
\useunder{\uline}{\ul}{}

\usepackage[breaklinks=true,bookmarks=false]{hyperref}

\iccvfinalcopy % *** Uncomment this line for the final submission

\def\iccvPaperID{5629} % *** Enter the ICCV Paper ID here

\ificcvfinal\fi

\begin{document}

%%%%%%%%% TITLE
\title{Hierarchical Visual Categories Modeling: A Joint Representation Learning and Density Estimation Framework for Out-of-Distribution Detection}

% \author{Jinglun Li, XinyuZhou, Pinxue Guo, Yixuan Sun, Yiwen Huang, Weifeng Ge, Wenqiang Zhang\\
% Fudan University\\
% {\tt\small jingli960423@gmail.com, zhouxinyu20@fudan.edu.cn, pxguo21@m.fudan.edu.cn,}\\
% {\tt\small 21210860014@m.fudan.edu.cn, 21210240056@m.fudan.edu.cn, weifeng.ge.ic@gmail.com,}\\
% {\tt\small wqzhang@fudan.edu.cn}

\author{Jinglun Li$^1$ \quad Xinyu Zhou$^2$ \quad Pinxue Guo$^1$ \quad Yixuan Sun$^1$ \quad Yiwen Huang$^2$\quad \\Weifeng Ge$^2\footnotemark[2]$ \quad Wenqiang Zhang$^{1,2}\footnotemark[2]$\\
$^1$Academy for Engineering and Technology, Fudan University\\
$^2$School of Computer Science, Fudan University$\quad$ \\
\tt\small jingli960423@gmail.com, zhouxinyu20@fudan.edu.cn, \\
\tt\small \{pxguo21, 21210860014, 21210240056\}@m.fudan.edu.cn, \\
\tt\small weifeng.ge.ic@gmail.com, wqzhang@fudan.edu.cn
}

\maketitle
\renewcommand{\thefootnote}{\fnsymbol{footnote}} %将脚注符号设置为fnsymbol类型，即特殊符号表示
\footnotetext[2]{ indicates corresponding authors.} %对应脚注[2]

\maketitle
\ificcvfinal\thispagestyle{empty}\fi

%%%%%%%%% ABSTRACT
\begin{abstract}
   Detecting out-of-distribution inputs for visual recognition models has become critical in safe deep learning. This paper proposes a novel hierarchical visual category modeling scheme to separate out-of-distribution data from in-distribution data through joint representation learning and statistical modeling. We learn a mixture of Gaussian models for each in-distribution category. There are many Gaussian mixture models to model different visual categories. With these Gaussian models, we design an in-distribution score function by aggregating multiple Mahalanobis-based metrics. We don't use any auxiliary outlier data as training samples, which may hurt the generalization ability of out-of-distribution detection algorithms. We split the ImageNet-1k dataset into ten folds randomly. We use one fold as the in-distribution dataset and the others as out-of-distribution datasets to evaluate the proposed method. We also conduct experiments on seven popular benchmarks, including CIFAR, iNaturalist, SUN, Places, Textures, ImageNet-O, and OpenImage-O. Extensive experiments indicate that the proposed method outperforms state-of-the-art algorithms clearly. Meanwhile, we find that our visual representation has a competitive performance when compared with features learned by classical methods. These results demonstrate that the proposed method hasn't weakened the discriminative ability of visual recognition models and keeps high efficiency in detecting out-of-distribution samples. 
\end{abstract}

%%%%%%%%% BODY TEXT
\section{Introduction}

\label{sec:introduction}

Modern deep neural networks have shown strong generalization ability when training and test data are from the same distribution~\cite{simonyan2014very, he2016deep, vaswani2017attention, devlin2018bert, liu2021swin}. However, encountering unexpected scenarios is inevitable in real-world applications. Thus assuring that training and test data share the same distribution becomes problematic. In applications like autonomous driving~\cite{blum2019fishyscapes,bogdoll2021description} and medical image analysis~\cite{qi2018adfor, guo2021cvad, shvetsova2021adin}, it is critical for models to identify inputs beyond their recognition capacity -- known as out-of-distribution (OOD) detection. OOD detection algorithms can enable the system to warn humans promptly in many safety-related scenarios. Moreover, it has become an important research topic in the research community of safe artificial intelligence~\cite{hendrycks17baseline, lee2018simpleuf, Hsu2020genodin, liu2020energy, winkens2020contrastive}.

{Many popular OOD detection methods aim to build probability models to describe training distributions~\cite{lee2018simpleuf,winkens2020contrastive,sun2022knnood,huang2021mos, Hsu2020genodin}. With these probability models, they built a score function that can calculate in-distribution scores for test samples. These in-distribution scores reflect whether these samples fall into a given distribution.
%\cite{hendrycks17baseline, lee2018simpleuf, Hsu2020genodin,liu2020energy,sun2022knnood,huang2021mos,haoqi2022vim}. 
Then the test sample can be evaluated by the score function to decide whether it is an OOD sample or not. 
Thus modeling features of in-distribution data become extremely important.
Previous works~\cite{lee2018simpleuf, haroush2021statistical, cao2022deep, fang2021learning, qiu2022latent} build a distribution over the whole training data. Since training images may come from various visual categories, the decision boundary between In-Distribution (InD) and OOD data becomes extremely complex. 
To solve this problem, subsequent studies~\cite{huang2021mos, chen2020boundary, han2022expanding, winkens2020contrastive} decomposed the whole dataset into several subgroups to simplify the decision boundary. Although representative algorithms like MOS~\cite{huang2021mos}, have gotten impressive performance in identifying OOD samples, they failed to detect near OOD samples. Because when different visual categories are grouped together, the OOD decision boundary will become even more uncertain. }

A typical framework for out-of-distribution detection involves two key steps: 1) learning a compact feature representation that can fit probability models easily; 2) modeling features of in-distribution data in complex distributions accurately. The above two problems are mutually connected because more compact features will make modeling the data distribution easier, and stronger probability modeling techniques will exploit fewer restrictions on representation learning. However, achieving the above goals is difficult, even when there are a lot of breakthroughs in deep learning. Because if training samples from the same category are too close in the feature space, it will usually lead to overfitting. Meanwhile, in-distribution samples may come from different visual categories that have large variations in appearance and semantic information, which makes modeling complex training distributions become challenging even for excellent statisticians. 

In this paper, we propose a new out-of-distribution detection framework, called {\em hierarchical visual category modeling}, to solve the above two issues simultaneously. We hold an assumption that given a training set that contains multiple visual categories, we can learn a probability model for every category independently. The out-of-distribution detection problem can be solved easily by aggregating probability models of known categories. Our motivation is that decomposing the whole dataset into subsets and modeling each category independently can avoid finding common characteristics shared by different categories. However, modeling an individual visual category is still challenging since classical supervised learning won't lead to compact feature representation. Thus, for each input sample, we need to force its feature representation to match the corresponding statistical model. That means we need to jointly conduct density estimation and representation learning. If we can jointly learn visual representations and optimize statistical models end-to-end, we can get good feature representations falling into distributions of the corresponding visual categories. Besides, we exploit knowledge distillation as done in~\cite{caron2021emerging} to learn robust feature representation. In this way, we can describe the complex training distribution with multiple Gaussian mixture models while not impairing the generalization ability of visual features.

In practice, to learn visual concepts that are in complex distributions, we build a Gaussian mixture model (GMM~\cite{rasmussen1999infinite}) for each visual category. Given input samples, we extract their deep features and project these features into a high-dimensional attribute space. Different from classical Gaussian mixture models that send the same input into $K$ different Gaussian models, we divide the attribute space into multiple groups and build a Gaussian model in each group independently. This strategy can give every attribute group a clear learning target and lead to better convergence. Experimental results indicate that this strategy works quite well. After the visual representation learning and statistical model parameters optimization, we can directly aggregate these statistical models to judge whether a test sample comes from the training distribution or not.  
To evaluate our OOD detector, we split ImageNet into ten folds randomly and select one of these splits as the training set and all other splits as the OOD dataset to conduct extensive tests. Experiments indicate that the proposed method has a strong ability to identify OOD samples.
We also evaluate our method on seven popular OOD benchmarks. Experimental results demonstrate that the proposed method not only can identify OOD samples efficiently but also improves the discriminative ability of learned visual representations. 

The contributions of this paper are summarized as follows:
% Contribution and novelty are not accurately positioned. The authors consider a fundamental and challenging problem (density estimation) but give a heuristic motivation from the perspective of out-of-distribution detection.
% Is it OK to rewrite the assumption at Line 177 as: “ Given a training set that contains multiple visual categories, if learn a probability model for every category independently, then we can obtain accurate density estimation (the key step 2 at Line 98). Therefore, the out-of-distribution detection problem can be solved much more easily by aggregating different probability models."
% The authors claim that this work proposes a novel out-of-distribution framework. Is the probability modeling novel? For out-of-distribution detection, is the probability modeling novel?
% determine the performance improvement of the model comes from which of the following modifications: (i) turning the post hoc mah method into fine-tuning version, (ii) introducing the multiple GMM models, (iii) proposing a novel loss function.
{\begin{itemize}
    %\item We introduce a new out-of-distribution detection scheme, called hierarchical visual category modeling to model multiple visual concepts that are in complex distributions. It provides a new perspective for out-of-distribution detection to build probability models for deep features.
    \item We introduce a new out-of-distribution detection scheme, called {\em hierarchical visual category modeling} to conduct joint representation learning and density estimation. It provides a new perspective for out-of-distribution detection to learn visual representation and probability models end-to-end.
    %\item We propose a novel training pipeline that can learn the parameters of density estimation models and visual representations efficiently. Specifically, we exploit multiple Gaussian mixture models to model every visual category and these Gaussian models can be learnt jointly with deep features.
    \item We exploit multiple Gaussian mixture models to model visual concepts in complex distributions. Visual attributes are divided into subgroups and modeled by different Gaussian components, which makes parameter learning much more efficient.
    
    %\item We exploit multiple Gaussian mixture models to model visual concepts in complex distributions. Visual attributes are divided into many sgroups and modeled by different Guassian components, which is proven to be .
    
    \item We conducted comprehensive experiments and ablation studies on popular benchmarks to investigate the effectiveness of the proposed method. Experiments demonstrated that our out-of-distribution detection models achieve better performance clearly when compared with previous methods.
\end{itemize} }

% The symbols in the manuscripts should correspond to those in Figure 1 for easy reading.
\begin{figure*}[!t]
  \centering
  \includegraphics[width=\linewidth]{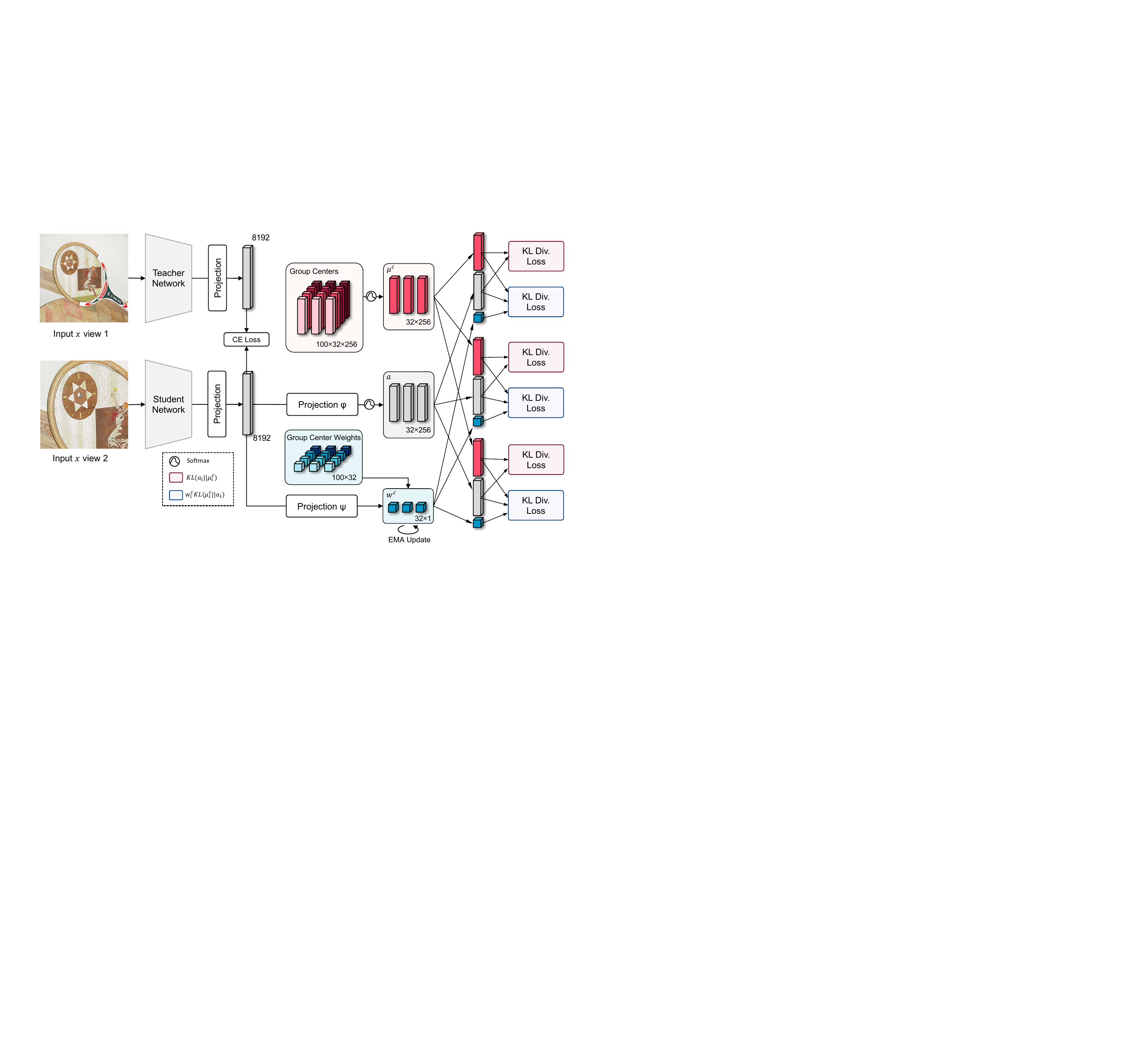}
  \caption{{Illustration of the training pipeline of hierarchical visual category modeling, written as HVCM. {In HVCM, we jointly learn the visual representation and parameters of probabilistic models. We get two different views of an input image and send the outputs into a knowledge distillation framework (DINO~\cite{caron2021emerging}). Image representations are projected into a high-dimensional attribute space. Then these attributes are divided into different groups and pass SoftMax functions to get attribute distributions. We match attributes in each group with stored attribute centers of the target visual category. The whole model is trained in an end-to-end manner.}}}
  % \vspace{-1.4em}
  \label{fig:framework}
\end{figure*}

\section{Related Work}

{\noindent\textbf{Out-Of-Distribution Detection.} Out-of-distribution detection aims to distinguish  out-of-distribution samples from in-distribution data. Numerous methods have been proposed for OOD detection. Maximum softmax probability (MSP)~\cite{hendrycks17baseline} has been recognized as a strong baseline by using the maximum score across all classes as an OOD score. ODIN~\cite{liang2018enhancing} improves MSP by perturbing the inputs and adjusting the logits via rescaling. Gaussian discriminant analysis has been employed in~\cite{lee2018simpleuf,winkens2020contrastive} to detect OOD samples. ReAct~\cite{sun2021react} uses rectified activation to reduce model overconfidence in OOD data. Shama {\em et al.} in~\cite{shama2019detecting} utilized Gram matrices to measure feature correlations for OOD detection. Bibas {\em et al.} in ~\cite{bibas2021single} proposed pNML regret to detect OOD samples with a single-layer neural network. The Generalized-ODIN approach~\cite{Hsu2020genodin} decomposes the confidence of class probability using a dividend/divisor structure to incorporate prior knowledge. Another promising line of work focuses on designing new learning objectives to train deep models from scratch. Leave-out Classifiers~\cite{vyas2018out} introduces a margin loss to ensure that InD and OOD samples are separated in the feature space. Lee {\em et al.} in~\cite{lee2018training} proposed a novel confidence loss to give lower confidence for OOD samples. There are also some methods~\cite{xiao2020likelihood, sensoy2020uncertainty, oberdiek2022uqgan} designed to conduct OOD detection based on generative models. Unlike these methods, our proposed method jointly conducts representation learning and density estimation to improve the OOD detection performance and keep the learned feature with strong discriminative ability.}

{\noindent\textbf{Density Estimation in Deep Learning.}
Density estimation tries to describe the probabilistic density distribution of observed data accurately and has been investigated in deep learning for a long time ~\cite{tabak2013family, dinh2016density, liu2021density}. In~\cite{an2015variational}, Chong {\em et al.} employed variational autoencoders for anomaly detection, where the reconstruction probability was used to compute the anomaly score of each sample. Papamakarios {\em et al}.~\cite{papamakarios2017masked} proposed a novel method for density estimation based on masked autoregressive flow. Zhou {\em et al.}~\cite{zhou2020gaussian} extended variational autoencoders by incorporating a mixture of Gaussians to model the latent space distribution, allowing for more flexible and expressive representations. Yang {\em et al.}~\cite{grover2018flow} combined Flow-based generative models with generative adversarial networks for density estimation and sample generation. Zhao {\em et al.}~\cite{dinh2016density} used discrete latent variables to conduct density estimation, which has been applied to many research topics in natural language processing and image processing. In this paper, we build probabilistic models that can be jointly learned with deep networks. We wish latent representations of the visual categories could easily be modeled by Gaussian mixture models even when complex distributions exist. Besides, we do not use post-hoc methods to conduct feature learning and density estimation separately. We aim to use the probabilistic model to guide the feature-learning process of deep networks.}

\section{Method} 

\subsection{Framework}\label{sec:3.1}
{
State-of-the-art methods~\cite{hendrycks17baseline, liang2018enhancing,lee2018simpleuf,huang2021mos, sun2022knnood,liu2020energy} typically assume all InD data follow the same distribution. However, such an assumption will lead to the decision boundary of an OOD detector being doped with some uncertain space. In this paper, we avoid modeling the whole InD dataset and only focus on independently modeling each object category's distribution. We propose hierarchical visual categories modeling to achieve this goal while maintaining high classification accuracy. In hierarchical visual categories modeling,  we first project image features into a high-dimensional attribute space (usually 8192 dimensions).
% The expression "attributes" is unclear and can have many meanings. Also in the literature on attribute learning, there exist several sub-notions of attributes. Please clarify. Is this more than a feature space?
These attributes can be grouped into multiple sub-visual concepts as components of an image category which can be easily modeled by Gaussian distributions.
Then combinations of multiple sub-visual concepts (abbreviated as sub-concepts) can be grouped to describe a more complex visual concept of an in-distribution category.
Since we define visual categories based on sub-visual concepts, simply modeling distributions of these sub-concepts and describing visual categories with these sub-concepts hierarchically can model complicated training distributions.
}

Formally, given a visual recognition model $f$, it maps an input image $\boldsymbol x$ with label $y$ to a high-dimensional feature vector $\boldsymbol{z} \in \mathbb{R}^q$. 
As described above, we project $\boldsymbol {z}$ into an attributes space $\mathcal{S}\in \mathbb{R}^{d}$ with a higher dimension and get an attribute description $\boldsymbol{a}$. 
Attributes in $\boldsymbol{a}$ can be grouped into multiple attribute subgroups $\{\boldsymbol{a}_i\}^G_{i=1}$ where $\boldsymbol{a}_i\in \mathcal{S}_i\subset\mathbb{R}^{d/G}$ and $i\in\{1, 2, \dots, G\}$. For images from a visual category $c$, we assume that attributes in their $i$-th attribute group follow a simple Gaussian distribution $\mathcal{N}(\mu^c_i, \Sigma^c_i)$, where $\mu^c_i$ and $\Sigma^c_i$ are the mean and variance respectively. Since attributes are divided into $G$ groups, we will have $G$ different Gaussian distributions for each visual category. The benefits of dividing the whole attribute space into several subspaces come from two folds: (1) modeling attribute distributions in these groups becomes easier; (2) assemble of distributions in these attribute groups can describe complex distributions, which will lead to more accurate decision boundaries. 
%In these subspaces, we can model each visual category using Gaussian distributions, and call. In this paper, we call these subspaces sub-visual concepts.
To combine all attributes to describe every visual category, our OOD detector learns weights $\{{w}^c_i\}^G_{i=1}$ of all attribute groups through exponential moving averages. 
Therefore, for each in-distribution class $c$, we can model the corresponding visual concept in a probability perspective with a mixture of Gaussian models:
\begin{equation}
    p\left ( \boldsymbol {x};c \right ) =\sum^G_{i=1}{w}^c_i\mathcal{N}(\boldsymbol {a}_i;\boldsymbol{\mu}^c_i, \boldsymbol{\Sigma}^c_i),
    \label{eq:class_pdf}
\end{equation}
where $w^c_i\in \mathbb{R}, \boldsymbol{\mu}^c_i\in \mathbb{R}^{d/G}, \boldsymbol{\Sigma}^a_i\in\mathbb{R}^{d/G\times d/G}$ and $\mathcal{N}(\cdot)$ means a normal distribution. We build a Gaussian mixture model for each class and get $C$ different Gaussian mixture models. Then, our proposed HVCM focuses on training deep neural networks to learn image features that follow the above distributions and parameters of these probability models jointly.

With the above Gaussian probability models, given a test sample $\boldsymbol {x}^{\prime}$, we define the score function $g(\boldsymbol{x}^{\prime}; \boldsymbol{w}^c, \boldsymbol{\mu}^c, \boldsymbol{\Sigma}^c)$ to measure whether it belongs to the $c$-th visual category using the learned probability density function. Here, $\boldsymbol{w}^c = \{{w}^c_i\}^G_{i=1}$, $\boldsymbol{\mu}^c =  \left \{ \boldsymbol{\mu}^c_i \right \} ^G_{i=1}$ and $\boldsymbol{\Sigma}^c =  \left \{ \boldsymbol{\Sigma}^c_i \right \} ^G_{i=1}$.
This score function can be used as a reliable metric to detect OOD samples:
\begin{equation}
    h(\boldsymbol{x}^{\prime})=\left\{
    \begin{aligned}
        &{\rm InD},  &{\rm if} \min_c g(\boldsymbol{x}^{\prime}; \boldsymbol{w}^c, \boldsymbol{\mu}^c, \boldsymbol{\Sigma}^c) \geq \gamma, \\
        &{\rm OOD},  &{\rm if} \min_c g(\boldsymbol{x}^{\prime}; \boldsymbol{w}^c, \boldsymbol{\mu}^c, \boldsymbol{\Sigma}^c) < \gamma,
    \end{aligned}
    \right.
    \label{eq:ood_detector}
\end{equation}
where $\gamma$ are thresholds to be determined in subsequent sections. Eq. (\ref{eq:ood_detector}) indicates that we use the minimal InD score among all $C$ categories to make the final decision.

Our framework is illustrated in Figure \ref{fig:framework}, where there are two steps: 
(1) jointly learn deep features that fit our probability models and parameters of these models; (2) calculate the InD score based on a set of Gaussian mixture models as the metric for out-of-distribution detection. In the following subsections, we give more details.

\subsection{Joint Visual Representation Learning and Parameter Optimization of Probability Models}
\label{sec:3.2}

To learn visual representations that follow Gaussian mixture models and keep their discriminative ability at the same time, we exploit the knowledge distillation framework DINO~\cite{caron2021emerging} to perform the joint learning. As shown in Figure~\ref{fig:framework}, for an image $\boldsymbol{x}$, we sample ten different views of $\boldsymbol{x}$ and send them into teacher and student branches simultaneously to perform self-distillation. 
% Sec 3.2.: Why are all dimensions already fixed in the method's description? Is it because this is what DINO provides you?
During knowledge distillation, we project {ResNet50~\cite{resnet2016} features in $2048$ dimensions into an attribute space with dimension $d (=8192)$.} Apart from the learning objective produced by self-distillation, we force the attributes of each class to follow a class-specific Gaussian mixture model. We divide the attributes $\boldsymbol{a} \in \mathbb{R}^{d}$ of $\boldsymbol{x}$ into $G$ groups to learn the Gaussian mixture model parameters. However, in practice, it's too hard to directly learn the mean and variance of Gaussian models. We follow He {\em et. al}~\cite{he2022hct} to learn attribute centers $\left \{ \boldsymbol{\mu}^c_i \right \} ^G_{i=1}$ of the $c$-th category (the image label $y$ is $c$). A linear projection layer is exploited to predict the weights $\{{w}^c_i(\boldsymbol{x})\}^G_{i=1}$ of $\boldsymbol{x}$ on all $G$ attribute groups. Our learning objective can be written as follows: 
% Do you normalize a_i and u^c_i to represent two distributions? Does this term represent the KL divergence between two normal distributions centered at a_i and u^c_i respectively?
% why did the researcher design a complex loss function in weighted KL divergence form in Eq.(3) rather than directly apply the standard maximum likelihood estimation method (i.e., in simpler L2 loss form) to fit the parameters in the GMM model? Is there a big difference between these two optimization methods? 
\begin{small}
\begin{align}
    \mathcal{L} = \mathcal{L}_{KD} + \alpha \sum^G_{i=1}{\rm KL}({\boldsymbol{a}}_i || \boldsymbol{\mu}^c_i)+ \beta \sum^T_{i=1}w^c_i\left ( \boldsymbol{x} \right ){\rm KL}(\boldsymbol{\mu}^c_i || {\boldsymbol{a}}_i)\label{eq:loss}
\end{align}
\end{small}
where $\mathcal{L}_{KD}$ stands for the cross entropy loss in self-distillation, KL stands for the Kullback-Leibler(KL) divergence, and $\alpha$ and $\beta$ are hyperparameters. The reason why we exploit two KL divergences is that experiments indicate that the term ${\rm KL}({\boldsymbol{a}}_i || \boldsymbol{\mu}^c_i)$ will favor learning attribute centers and the term ${\rm KL}(\boldsymbol{\mu}^c_i || {\boldsymbol{a}}_i)$ is more suitable to learn image attribute descriptions and group weights. Note that softmax operations normalize attributes and their learnable centers in each group before calculating learning objectives.   

With the learning objective in Eq. (\ref{eq:loss}), student network parameters $\theta_s$, and weights $\{{w}^c_i\}^G_{i=1}$ and attribute centers $\left \{ \boldsymbol{\mu}^c_i \right \} ^G_{i=1}$ of all groups are learnt at the same time. They are updated using the following equations:
\begin{align}
    \theta^{t+1}_s &= \theta^{t}_s - \gamma _1 \frac{\partial \mathcal{L}}{\partial \theta^{t}_s}, \label{eq:update_param} \\
    \boldsymbol{\mu}^{c,t+1}_{i} &= \boldsymbol{\mu}^{c,t}_{i} - \gamma _2 \frac{\partial \mathcal{L}}{\partial \boldsymbol{\mu}^{c,t}_{i}}, \label{eq:update_centers}\\
    \boldsymbol{w}^{c,t+1}_{i} &= (1-\gamma_3) \boldsymbol{w}^{c,t}_{i} + \gamma _3 {w}^c_i(\boldsymbol{x}),\label{eq:update_weights}
\end{align}
where $\gamma_1$, $\gamma_2$ and $\gamma_3$ are the learning rates. $\gamma_1, \gamma_2$ control the gradient update speed, while $\gamma_3$ controls the updating speed of the group weights. Following He {\em et. al}~\cite{he2022hct}, $\theta_s$, $\left \{ \boldsymbol{w}^c \right \} _{c=1}^C$ and $\left \{ \boldsymbol{\mu}^c \right \} _{c=1}^C$ are initialized with Gaussian noises. We use Adam optimizer~\cite{kingma2014adam} with momentum to update both $\theta_s$ and $\left \{ \boldsymbol{\mu}^c \right \} _{c=1}^C$. While the attribute weights are learned through exponential moving averages~\cite{he2022masked}.

\begin{figure}[!t]
  \centering
    \includegraphics[width=\linewidth]{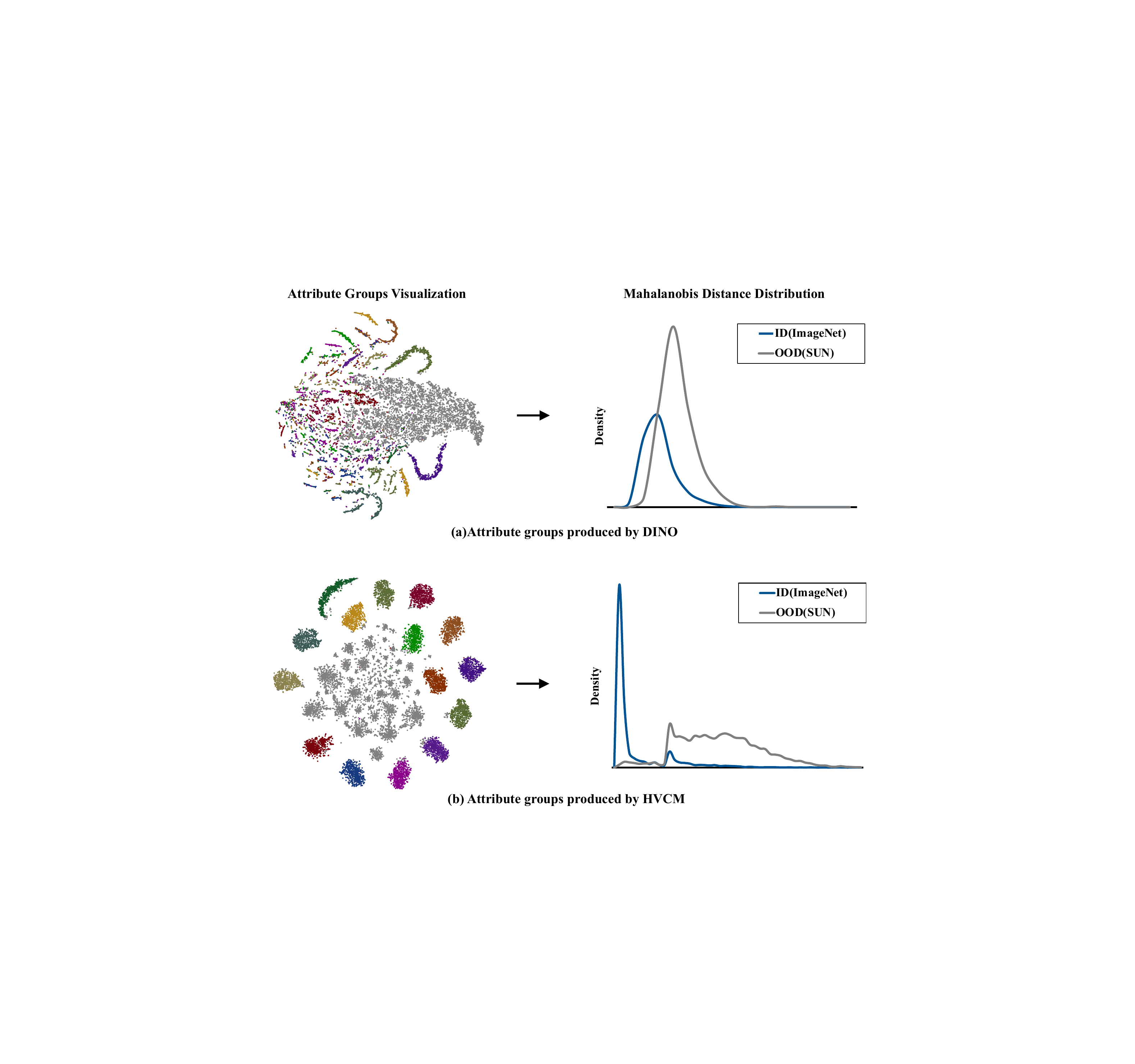}
    \caption{Illustration of attribute group visualization and Mahalanobis distance distribution. Both attribute groups are visualized by t-SNE~\cite{van2008visualizing}. The colors encode different in-distribution data(ImageNet), and out-of-distribution features(SUN) marked as gray points. Models are trained on ResNet-50~\cite{he2016deep} using DINO(a) and HVCM(b).}
    % \vspace{-1em}
    \label{fig:density}
\end{figure}

\subsection{Out-of-Distribution Detection based on Probability Models in Hierarchy}
When the training of our hierarchical probability model converges, we will get visual attributes that follow learned Gaussian distributions hierarchically for in-distribution samples. Meanwhile, we also get weights and centers of all attribute groups for each class. 
% Is there any convergence guarantee for the training of the hierarchical probability model? (Line 408) If the training can converge, is it suitable to use the converged value of the mean attributes as group centers?
% Can the given approach be understood 1) as assuming a block-diagonal covariance structure? 2)Learning a model that provides outputs that can be modeled by 1). If yes, I believe this could be mentioned as an additional entry hook.
{However, the mean attributes are updated frequently during the training and are thus not suitable to be used as group centers. So we need to recalculate the attribute centers for each visual category.}  Given an in-distribution visual category $c$, we estimate the mean vector and covariation matrix of the $i$-th attribute group:
\begin{align}
    {\boldsymbol {\mu}}^c_i&= \frac{1}{N_c}\sum_{m=1}^{N_c}\boldsymbol{a}^m_i\\
    \boldsymbol {\Sigma}^c_i&= \frac{1}{N_c-1}\sum_{m=1}^{N_c}(\boldsymbol{a}^m_i-\boldsymbol{\mu}^c_i)(\boldsymbol{a}^m_i-\boldsymbol{\mu}^c_i)^\top, 
    \label{eq:sigma}
\end{align}
where $N_c$ notes the number of samples in the  $c$-th class, and $\boldsymbol{a}^m_i$ is the sub-attribute vector of the $m$-th sample. 
%The reason why we don't follow \cref{eq:sigma} to estimate the mean attributes is because we find that these learnt attribute centers work much better than simply averaging as described in Section XXX.
With these weights, means, and covariances, we can describe each visual category in the hierarchy accurately. We try to use the probability density function in Eq. (\ref{eq:class_pdf}) as the in-distribution function. However, we will encounter the problem of numerical overflow when calculating the determinants of covariance matrices as~\cite{overflow}. Instead, we compute the Mahalanobis distance between the $i$-th sub-visual attributes $\boldsymbol{a}^{\prime}_i$ of a test sample $\boldsymbol{x}^{\prime}$ and the corresponding attribute center $\boldsymbol{\mu}^c_i$ to measure the likelihood of these attributes belongs to the target category:
\begin{align}
    M^c_i(\boldsymbol{x}) = -(\boldsymbol{a}^{\prime}_i-\boldsymbol{\mu}^c_i)^\top \left ( \boldsymbol {\Sigma}^c_i \right )^{-1}  (\boldsymbol{a}^{\prime}_i-\boldsymbol{\mu}^c_i) .
\end{align}
% According to Eq. (1), the input x with label c is generated from the i-th sub-concepts with probability w^c_i. Could you explain more about the class-level score function in (10)? Why does this score use weighted sum rather than max?
{The above equation gives the in-distribution measure for one attribute group. While, for every category, we have multiple attribute groups and need to assemble related in-distribution measures to get the class-level in-distribution score. Since we have gotten the weights of attribute groups for each category, we can easily assemble them and get the class-level score function:}  
\begin{align}
    g(\boldsymbol{x}^{\prime}; \boldsymbol{w}^c, \boldsymbol{\mu}^c, \boldsymbol{\Sigma}^c) = \sum_{i=1}^{G} w^c_i  M^c_i(\boldsymbol{x}^{\prime}).
    \label{eq:class_score}
\end{align}
With this score function, we can easily get the in-distribution score of a test sample on each visual category. Since there are $C$ categories in the whole in-distribution dataset, we get the maximal in-distribution score across different visual categories as the in-distribution score on the whole dataset:
\begin{align}
    g(\boldsymbol{x}^{\prime}) = \max_c  g(\boldsymbol{x}^{\prime}; \boldsymbol{w}^c, \boldsymbol{\mu}^c, \boldsymbol{\Sigma}^c).
\end{align}
A high in-distribution score $g(\boldsymbol{x}^{\prime})$ indicates that the semantic attributes of a test sample lie very close to one or multiple in-distribution visual categories(as shown in Figure~\ref{fig:density}). On the contrary, if a sample does not belong to the previously modeled categories, it will get a low in-distribution score. We follow Eq. (\ref{eq:ood_detector}) to set thresholds to judge whether a sample is an out-of-distribution sample. In the experiment section, we discuss how to set these thresholds.

\begin{table*}
  \setlength\belowdisplayskip{0.3cm}
\caption{OOD detection performance comparison of HVCM and existing methods.  All comparison methods rely on ResNet-50 trained with cross-entropy loss. * indicates that the method is fine-tuned on InD data. $\uparrow$ indicates larger values are better, and $\downarrow$ is the opposite. \textbf{Bold} numbers are superior results. All values are percentages. 
  }
\vspace{0.5em}

\label{tab:mresults1}
\tabcolsep=0.18cm
\footnotesize
\begin{tabular*}{\linewidth}{@{}lccccccccccc@{}}
\toprule
\multicolumn{1}{c}{\multirow{3}{*}{\textbf{Method}}} & \multicolumn{8}{c}{\textbf{OOD Datasets}}                                                                                                                 & \multicolumn{2}{c}{\multirow{2}{*}{\textbf{Average}}} & \multirow{3}{*}{\textbf{InD Acc}} \\ \cmidrule(lr){2-9}
\multicolumn{1}{c}{}                                 & \multicolumn{2}{c}{\textbf{iNaturalist}} & \multicolumn{2}{c}{\textbf{SUN}} & \multicolumn{2}{c}{\textbf{Places}} & \multicolumn{2}{c}{\textbf{Textures}} & \multicolumn{2}{c}{}                                  &                                  \\
\multicolumn{1}{c}{}                                 & \textbf{FPR95↓}         & \textbf{AUROC↑}        & \textbf{FPR95↓}     & \textbf{AUROC↑}    & \textbf{FPR95↓}      & \textbf{AUROC↑}      & \textbf{FPR95↓}       & \textbf{AUROC↑}       & \textbf{FPR95↓}                & \textbf{AUROC↑}              &                                  \\ \midrule
MSP~\cite{hendrycks17baseline}  & 68.12               & 87.48              & 58.59           & 89.76          & 59.53            & 89.18            & 72.66             & 82.71             & 64.73                     & 87.28                   & \multirow{7}{*}{85.74}           \\
ODIN~\cite{liang2018enhancing}  & 55.78               & 85.92              & 60.47           & 83.83          & 60.94            & 88.06            & 64.06             & 81.28             & 60.31                      & 84.77                    &                                  \\
Maha~\cite{lee2018simpleuf}     & 97.00               & 55.12              & 98.80           & 51.93          & 97.20            & 50.52            & \textbf{20.00}    & \textbf{94.99}    & 78.25                      & 63.14                    &                                  \\
Energy~\cite{liu2020energy}     & 58.91               & 86.45              & 27.03           & 93.44          & 38.75            & 91.51            & 56.25             & 87.26             & 45.24                      & 89.67                    &                                  \\
GODIN*~\cite{Hsu2020genodin}    & 72.00               & 79.86              & 60.09           & 84.58          & 64.94            & 82.30            & 39.50             & 89.28             & 59.13                      & 84.01                    &                                  \\
MOS*~\cite{huang2021mos}        & 52.94               & 91.41              & 67.78           & 86.82          & 71.31            & 84.38            & 73.65             & 80.56             & 66.42                      & 85.79                    &                                  \\
ReAct~\cite{sun2021react}        & 58.48               & 82.60              & 78.18           & 69.11          & 86.33            & 59.84            & 50.53             & 87.03             & 68.38                      & 74.65                    &                                  \\
\midrule
\textbf{HVCM(Ours)}                               & \textbf{21.56}      & \textbf{92.19}     & \textbf{17.20}  & \textbf{94.44} & \textbf{19.98}   & \textbf{93.62}   & 29.22             & 90.68             & \textbf{21.99}            & \textbf{92.73}           & \textbf{88.28}                   \\ \bottomrule
\end{tabular*}
% \vspace{-1em}
\end{table*}

\section{Experiments}
\subsection{Experimental Setup}
\label{sec:setup}
\noindent\textbf{In-distribution Datasets.} We use ImageNet-1K~\cite{russakovsky2015imagenet} and CIFAR10~\cite{krizhevsky2009learning} as our in-distribution datasets. ImageNet-1K is a large-scale visual recognition dataset containing 1000 object categories and 1281167 images. We split it into 10 folds randomly and ensured each fold contain 100 object categories. Since our computation resources are limited, we randomly select one fold as the in-distribution dataset. There other nine folds are used as OOD datasets as other popular benchmarks to test the performance of the proposed method in detecting near OOD samples. For CIFAR10~\cite{krizhevsky2009learning}, there are 60000 color images in 10 classes. We conduct OOD algorithm evaluation as previous methods~\cite{liang2018enhancing,de2000mahalanobis,sun2022knnood,liu2020energy, sun2021react, wei2022mitigating}.

\noindent\textbf{Out-of-distribution Dataset.} 
On ImageNet, we follow Huang {\em et al.}~\cite{huang2021mos} to test our methods and use Texture~\cite{text2014}, iNaturalist~\cite{van2018inaturalist}, Places365~\cite{places2018}, and SUN~\cite{sun2010}) as OOD test sets. To further explore the limitation of our approach, we evaluate our method on another two OOD datasets, OpenImage-O~\cite{krasin2017openimages} and ImageNet-O~\cite{hendrycks2021nae}. For CIFAR10, as in~\cite{sun2022knnood,liu2020energy, sun2021react, wei2022mitigating}, we selected eight widely-used datasets, including Texture~\cite{text2014}, SVHN~\cite{netzer2011reading}, Places365~\cite{places2018}, iSUN~\cite{xu2015turkergaze}, LSUN-Crop~\cite{yu2015lsun}, LSUN-Resize~\cite{yu2015lsun},ImageNet-Resize~\cite{russakovsky2015imagenet}, and ImageNet-Fix~\cite{russakovsky2015imagenet} as our test sets. Besides, to test the ability of HVCM to identify near OOD datasets, we use the remaining nine ImageNet subsets as the OOD test sets. {\em Note that since our evaluation on ImageNet differs from previous methods~\cite{huang2021mos,sun2021react}, we implement these algorithms with open source provided by authors and follow standard experimental settings.}

\noindent\textbf{Evaluation Metrics.} We employ the commonly used metrics in OOD detection~\cite{huang2021mos, sun2021react} to evaluate our approach, including AUROC, FPR95 and InD Acc. AUROC stands for the area under the receiver operating characteristic curve, FPR95 is short for TPR@FPR95 and represents the false positive rate when the true positive rate is 95\%, and InD Acc is the classification accuracy of in-distribution data.

% The settings of some critical hyperparameters need to be clarified. For example, multiple stable learning rates appeared in Eq. (4)-(6), but only one learning rate is given in the implementation part. The loss function weights in Eq.(3) should be given in the implementation part. Since many hyperparameters are introduced in this study

\begin{table}[!t]
% \vspace{-1em}
\caption{Evaluation on more challenging detection tasks. * indicates that the method is fine-tuned on InD data. $\uparrow$ indicates larger values are better, and $\downarrow$ is the opposite. \textbf{Bold} numbers are superior results. All values are percentages.}
\vspace{0.5em}
\label{tab:mresults2}
\tabcolsep=0.2cm
\footnotesize
\begin{tabular*}{\linewidth}{lcccc}
\toprule
\multicolumn{1}{c}{\multirow{2}{*}{\textbf{Method}}} & \multicolumn{2}{c}{\textbf{ImageNet-O}} & \multicolumn{2}{c}{\textbf{OpenImage-O}} \\
\multicolumn{1}{c}{}                                 & \textbf{FPR95↓}      & \textbf{AUROC↑}    & \textbf{FPR95↓}      & \textbf{AUROC↑}   \\ \hline
MSP~\cite{hendrycks17baseline}       & 72.50                             & 83.33                               & 89.11                             & 57.99                               \\
ODIN~\cite{liang2018enhancing}       & 73.44                             & 71.03                               & 64.06                             & 79.88                               \\
Maha~\cite{lee2018simpleuf}          & 54.40                             & 79.30                               & 59.20                             & 77.62                               \\
Energy*~\cite{liu2020energy}          & 66.56                             & 81.06                               & 62.34                             & 84.61                               \\
GODIN*~\cite{Hsu2020genodin}         & 71.55                             & 79.89                               & 73.57                             & 77.27                               \\
MOS*~\cite{huang2021mos}             & 87.40                             & 64.87                               & 66.78                             & 81.42                               \\
ReAct~\cite{sun2021react}             & 87.05                             & 64.15                               & 84.17                             & 64.30                               \\
\textbf{HVCM(Ours)}              & \textbf{42.86}                    & \textbf{86.72}                      & \textbf{28.58}                     & \textbf{90.26}                      \\ \bottomrule
\end{tabular*}
% \vspace{-1.7em}
\end{table}

\noindent\textbf{Training Details.}{ We utilize ResNet-50~\cite{resnet2016} as the feature backbone for ImageNet and the dimension of the attribute space is set to 8192. The training is finished in 300 epochs. On CIFAR10, we use ResNet-18~\cite{resnet2016} as our feature backbone, and the dimension of the attribute space is set to 1024. The training on CIFAR10 is finished in 200 epochs. The number of attribute groups is set to 32, 
and $\alpha$, $\beta$, $\gamma_1, \gamma_2$, and $ \gamma_3$ are set to 1, 0.1, 1, 1 and $1\times 10^{-4}$, respectively. 
Updating $\mu$ too fast can lead to the oscillation of group centers, negatively influencing the precision of probabilistic modeling. So we utilize a smaller $\beta$ than $\alpha$ to update $\mu$. All the hyperparameters are tuned according to the experimental results.
We employ SGD with a momentum of 0.9, an initial learning rate of 0.1, and a batch size of 128. The learning rate is reduced by a factor of 10 at 50\% and 75\% of the total training epochs. We train all backbones from scratch using random initialization. All experiments are performed using PyTorch~\cite{paszke2019pytorch} with default parameters on four NVIDIA GeForce RTX 3090.}

\begin{table*}[!t]
\tabcolsep=0.12cm
\scriptsize
\caption{Comparison of OOD detection performance of HVCM and existing methods on CIFAR10 dataset. All compared methods use ResNet-18 trained with cross-entropy loss except our proposed method, which uses HVCMLoss. The performance is evaluated based on AUROC (A) and FPR95 (F). ↑ indicates larger values are better and ↓ indicates the opposite. \textbf{Bold} numbers indicate superior results. All values are expressed in percentages.}
\vspace{0.5em}

\label{tab:mresults3}

\begin{tabular*}{\linewidth}{@{}llcccccccccccccccccc@{}}
\toprule
\multicolumn{2}{c}{\multirow{3}{*}{\textbf{Method}}} & \multicolumn{16}{c}{\textbf{OOD Dataset}}                                                                                                                                                                                                                                                                                         & \multicolumn{2}{c}{\multirow{2}{*}{\textbf{Average}}}             \\ \cmidrule(lr){3-18}
\multicolumn{2}{c}{}                                 & \multicolumn{2}{c}{\textbf{Texture}} & \multicolumn{2}{c}{\textbf{SVHN}}    & \multicolumn{2}{c}{\textbf{Places365}} & \multicolumn{2}{c}{\textbf{iSUN}} & \multicolumn{2}{c}{\textbf{LSUN(C)}} & \multicolumn{2}{c}{\textbf{LSUN(R)}} & \multicolumn{2}{c}{\textbf{ImageNet(R)}} & \multicolumn{2}{c}{\textbf{ImageNet(F)}} & \multicolumn{2}{c}{}                                              \\
\multicolumn{2}{c}{}                                 & \textbf{F↓}         & \textbf{A↑}    & \textbf{F↓}   & \textbf{A↑}          & \textbf{F↓}       & \textbf{A↑}        & \textbf{F↓}     & \textbf{A↑}     & \textbf{F↓}      & \textbf{A↑}       & \textbf{F↓}      & \textbf{A↑}       & \textbf{F↓}          & \textbf{A↑}           & \textbf{F↓}         & \textbf{A↑}         & \multicolumn{1}{c}{\textbf{F↓}} & \multicolumn{1}{c}{\textbf{A↑}} \\ \midrule
\multirow{4}{*}{CELoss}     & MSP                    & 56.47               & 90.20          & 58.40         & 90.56                & 51.86             & 91.98              & 53.84           & 91.64           & 46.22            & 92.74             & 49.10            & 92.48             & 59.65                & 89.35                 & 59.90               & 89.40               & 54.43                           & 91.04                           \\
                            & ODIN                   & 40.37               & 91.98          & 27.82         & 93.28                & 31.80             & 94.23              & 17.59           & 96.60           & 25.23            & 95.31             & 14.74            & 97.21             & 27.18                & 94.65                 & 45.26               & 90.62               & 28.75                           & 94.24                           \\
                            & Gram                   & 10.81               & 97.73          & 2.58          & 99.39                & 30.15             & 93.37              & 1.11            & {\ul 99.76}     & 13.91            & 97.03             & 0.52             & 99.86             & 1.61                 & 98.38                 & 66.94               & 82.02               & 15.95                           & 95.94                           \\
                            & Gram+pNML              & {\ul \textbf{7.18}} & {\ul 98.50}    & {\ul 1.63}    & \textbf{99.60}       & {\ul 23.21}       & {\ul 95.13}        & {\ul 0.83}      & \textbf{99.80}  & {\ul 9.42}       & {\ul 98.00}       & \textbf{0.42}    & {\ul 99.88}       & \textbf{1.24}        & 98.76                 & 57.90               & 85.53               & 12.73                           & 96.90                           \\ \midrule
\textbf{HVCMLoss}           & \textbf{HVCM(ours)}    & \textbf{1.88}       & \textbf{99.31} & \textbf{1.32} & {\ul \textbf{99.47}} & \textbf{0.95}     & \textbf{99.54}     & \textbf{0.65}   & 99.71           & \textbf{0.77}    & \textbf{99.66}    & {\ul 0.51}       & \textbf{99.96}    & 1.80                 & \textbf{99.20}        & \textbf{4.77}       & \textbf{98.19}      & \textbf{1.58}                   & \textbf{99.38}                  \\ \bottomrule
\end{tabular*}
\end{table*}

\subsection{Comparison with State-of-the-Art Algorithms}

\noindent\textbf{Standard evaluation on ImageNet.}
We compare our HVCM with seven popular OOD detection methods, including MSP~\cite{hendrycks17baseline}, ODIN~\cite{liang2018enhancing}, GODIN~\cite{Hsu2020genodin}, Maha ~\cite{lee2018simpleuf}, Energy~\cite{liu2020energy}, MOS~\cite{huang2021mos}, and ReAct~\cite{sun2021react}. For datasets that describe objects or scenes, such as SUN, Places, and iNaturalist, HVCM achieves better AUROC and FPR95 metrics. When we summarize the results of all four datasets, HVCM achieves 21.99\% on FPR95 and 92.73\% on AUROC, which outperforms the previous best method Energy~\cite{liu2020energy} by 23.25\% and 3.06\%. This is a significant improvement, which demonstrates that end-to-end training is very important to get good results. 
% When compared with Maha~\cite{lee2018simpleuf}, we get inferior performance both on FPR95 and AUROC. It indicates that the proposed method is weak for describing textures. We think Textures contain many repeated patterns which are different from general object recognition. Maha~\cite{lee2018simpleuf} used features in both intermediate and deep layers of neural networks, while we only employed features in the last layer of the network. 
When compared to Maha~\cite{lee2018simpleuf}, our proposed method exhibits inferior performance in terms of both FPR95 and AUROC. This observation suggests that our method is less effective in describing textures. We attribute this limitation to the fact that textures often encompass numerous repeated patterns, which differ from the characteristics required for general object recognition. It is important to note that our method only utilizes features from the last layer of the network, while Maha~\cite{lee2018simpleuf} leverages features from both intermediate and deep network layers.
However, although our results are worse than Maha~\cite{lee2018simpleuf}, we are better than all the other methods. This still indicates that the proposed method is very robust in identifying different types of outliers. Furthermore, we build a cosine classifier with the learned attribute centers for image classification. We got 88.28\% accuracy which is 2.57\% higher than our supervised learning baseline. This is solid evidence to show that our proposed method can model the InD data accurately and ensure the learned features keep high discriminative ability simultaneously.

\begin{figure*}[!t]
  \centering
    \includegraphics[width=\linewidth]{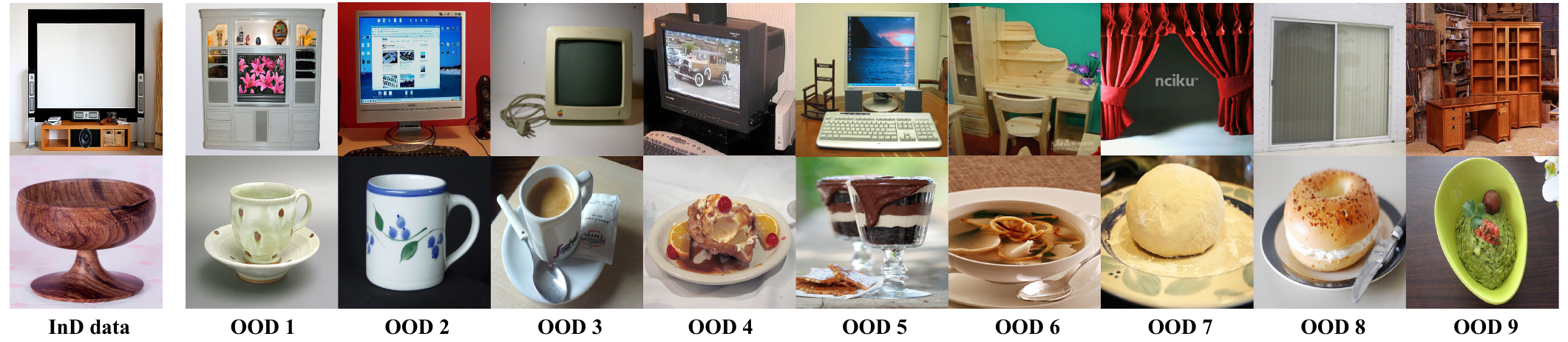}
    \caption{Each dataset exhibits displayed two categories of images. The leftmost samples belong to the InD dataset, while the categories on the right correspond to nine OOD datasets arranged in ascending order of distance. It can be observed that the gap between the OOD samples and InD samples gradually widens as the distance increases.}
    \label{fig:sample}
    % \vspace{-1em}
\end{figure*} 

{\setlength{\parindent}{0cm}
\textbf{Standard evaluation on CIFAR10.}
% (a) some of the more cutting-edge methods in recent years have not been compared, e.g., Gram [2], pNML[3], etc. (b) lack of testing on multiple different backbone networks, which is very common in many previous studies. c) The article declares the proposed method can solve the near OOD sample detection problem. I suggest the author conduct an experiment and comparison on one of the most common near OOD detection settings, i.e., the CIFAR setting in which CIFAR10 and CIFAR100 are used as ID and OOD, respectively. (d) the DINO training scheme does not seem to be related to the main contribution claimed by this paper. What about removing the DINO training scheme and directly tuning the pre-trained classifier like traditional OOD detectors? Such a result needs to be included in the ablation study part. 
% In how far is DINO a key ingredient? Can this be demonstrated by another experiment?
{We performed a more conventional OOD detection task on CIFAR10. This experiment aims twofold: firstly, to demonstrate that our method is not dependent on self-distillation (DINO~\cite{caron2021emerging}), and secondly, to validate the effectiveness and robustness of our HVCM. Table~\ref{tab:mresults3} compares our method with several classic and top-performing algorithms. All comparison methods use ResNet-18 as the main backbone network and are trained with cross-entropy loss, while our method uses only the loss function in Eq. (\ref{eq:loss}).
As shown in Table~\ref{tab:mresults3}, the proposed method outperforms the previous best methods Gram~\cite{shama2019detecting} and pNML~\cite{bibas2021single} on both average FPR95 and AUROC obviously. These results demonstrate the proposed method can perform well even on small datasets. Meanwhile, there is no self-distillation exploited, which indicates our joint representation learning and statistical modeling is independent of self-supervised learning algorithms~\cite{caron2021emerging}.}}

\noindent\textbf{Results on more challenging OOD datasets.} To overcome the limitations of current OOD benchmarks~\cite{haoqi2022vim} and evaluate the robustness of our approach against adversarial attacks, we conducted experiments on two challenging datasets, namely OpenImage-O~\cite{krasin2017openimages} and ImageNet-O~\cite{hendrycks2021nae}. As shown in Table \ref{tab:mresults2}, HVCM achieves the highest AUROC and lowest FPR95 among all methods on the OpenImage-O dataset. Although ImageNet-O contains adversarial examples and is more challenging, HVCM still outperforms other methods on this dataset.

\begin{figure}[H]
  \centering
    \scalebox{1.0}{
    \includegraphics[width=\linewidth]{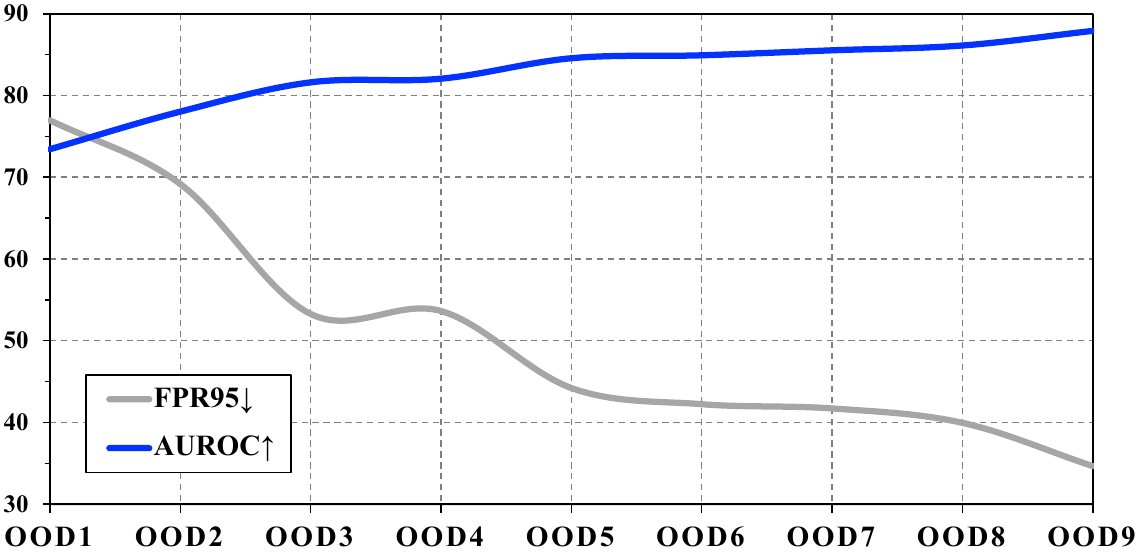}
    }
    \caption{HVCM performance comparison as the increasing distances between InD and OOD data.}
    \label{fig:near2far}
% \vspace{-1.4em}
\end{figure}

\noindent\textbf{Results on near-to-far OOD datasets.}
To investigate the ability of the proposed method to detect near OOD samples, we construct 9 OOD test sets with the remaining ImageNet images. We rank the semantic distance between the remaining 900 visual categories with the 100 categories in the InD dataset. We use the average cosine distance as the measure and construct 9 different OOD test sets. Details are introduced in the supplementary material. For convenience, we denote these datasets from OOD 1 to OOD 9.
The experimental results are depicted in Figure \ref{fig:near2far}, and several samples are displayed in Figure \ref{fig:sample}. We can find that the proposed method achieves good AUROC even when the test set is very close to the InD dataset. When the test sets become farther, FPR95 decreases quickly, which indicates the proposed method is very sensitive to the semantic distances of OOD datasets.

\subsection{Ablation Study}
\noindent\textbf{Number of attribute groups.}
% Sec. 4.3: why didn't the authors also consider greater values of G?
We varied the number of attribute groups from 8 to 32 to analyze the components in Gaussian mixture models. In Table \ref{tab:numsubcon}, we find a positive correlation between the number of attribute groups and model performance, with the best performance achieved when $G$ is set to 32. We also try to set $G$ to bigger numbers and list the results in the supplementary material. However, more attribute groups will lead to bigger correlation matrices when computing the InD score. Thus, we finally set $G$ to 32 to balance performance and inference speed.

\begin{table}[H]
% \vspace{-2em}
\caption{The performance of HVCM with varying numbers of group centers $G$. Results are averaged across four standard OOD datasets, consistent with the main results.}
\vspace{0.5em}
\label{tab:numsubcon}
\centering
\footnotesize
\tabcolsep=0.6cm
\scalebox{0.9}{
\begin{tabular*}{\linewidth}{@{}ccc@{}}
\toprule
\multicolumn{1}{l}{\textbf{Concepts Number}} & \textbf{FPR95↓} & \textbf{AUROC↑} \\ \midrule
$G$=8                              & 24.37         & 91.73           \\
$G$=16                             & 23.07         & 92.24           \\
$G$=\textbf{32}                    & \textbf{21.99}         & \textbf{92.73}           \\ \bottomrule
\end{tabular*}
}
% \vspace{-2em}
\end{table}

{\setlength{\parindent}{0cm}
\textbf{Choice of learning objectives.}
{The first row of Table \ref{tab:ablation} investigates how the choice of the learning objective influence the performance of HVCM. We test three learning objectives, including the L2 loss, JS divergence loss, and KL divergence loss. The results show that compared with the L2 loss and KL divergence loss, the JS divergence loss achieves the lowest FPR95, demonstrating its superiority for statistical modeling. We attribute this to the fact that the group centers and features need to learn from each other, and the JS divergence loss is symmetrical in enclosing them.}

}

%{\setlength{\parindent}{0cm}\textbf{Different training strategies.}We compare different training strategies for HVCM and present multiple pieces of evidence to demonstrate the advantages of HVCM. As shown in Figure \ref{fig:density} (a), the self-distillation technique can produce more unique features than regular supervised learning, but these features are dispersed in the feature space, which limits the performance of OOD detection. However, HVCM effectively mitigates this issue, as shown in Figure \ref{fig:density} (b), by producing more compact features that make the in-distribution and out-of-distribution Mahalanobis distance separable. In addition to visualizations, we quantitatively evaluate the performance of each strategy in the middle row of Table \ref{tab:ablation}. The results demonstrate that HVCM with self-distillation outperforms other strategies by a considerable margin.}

\begin{table}[H]
\caption{A set of ablation results about HVCM. The top row investigates the effect of using MSE, KL divergence, or JS divergence on the model performance; the bottle row compares the performance of different OOD detection methods. Results are averaged across four standard OOD datasets following the main results.}
\vspace{0.5em}
\label{tab:ablation}

\tabcolsep=0.45cm
\centering
\footnotesize
\scalebox{0.9}{
\begin{tabular*}{\linewidth}{@{}ll|cc@{}}
\toprule
\multicolumn{2}{c}{\textbf{Strategy   Ablation}}     & \textbf{FPR95↓}  & \textbf{AUROC↑} \\ \midrule
\multirow{2}{*}{Learning Objective}      & L2                   & 23.59                 & 92.26                \\
& KL             & 22.99          & 92.58           \\
                                    & \textbf{JS}    & \textbf{21.99} & \textbf{92.73}  \\ \midrule
\multirow{3}{*}{InD Distance}       & Cosine         & 66.37          & 84.06           \\
                                    & Linear         & 36.77          & 86.59           \\
                                    & \textbf{Maha}  & \textbf{21.99} & \textbf{92.73}  \\ \bottomrule
\end{tabular*}}
% \vspace{-1em}
\end{table}

{
{\setlength{\parindent}{0cm}
\textbf{Different InD distance metrics.} In Table \ref{tab:ablation}, we also compared our Maha metric with two different InD distance metrics. The cosine distance metric directly measures the distance by calculating the cosine similarity between the input feature and the mean of Gaussian distribution models. The linear distance metric is used to calculate the distance with trainable linear layers. The results show that our Maha metric is an effective metric compared with its counterparts. We attribute this to the Maha distance space can better fit the training distribution in realistic scenes.
% including cosine distance. The cosine similarity strategy, which does not require covariance, performs poorly. To implement the linear strategy, we use Mahalanobis distances computed from one sample with 3200 Gaussian models of all InD classes as the input. The goal is to convert OOD detection into a binary classification problem using deep learning. Unfortunately, this deep learning approach yields bad results. Our experimental results demonstrate that the most effective OOD detection strategy is to hierarchically compute the Mahalanobis distances and find the closest InD categories.
}}

\begin{figure}[!h]
  \centering
  \scalebox{0.95}{
    \includegraphics[width=\linewidth]{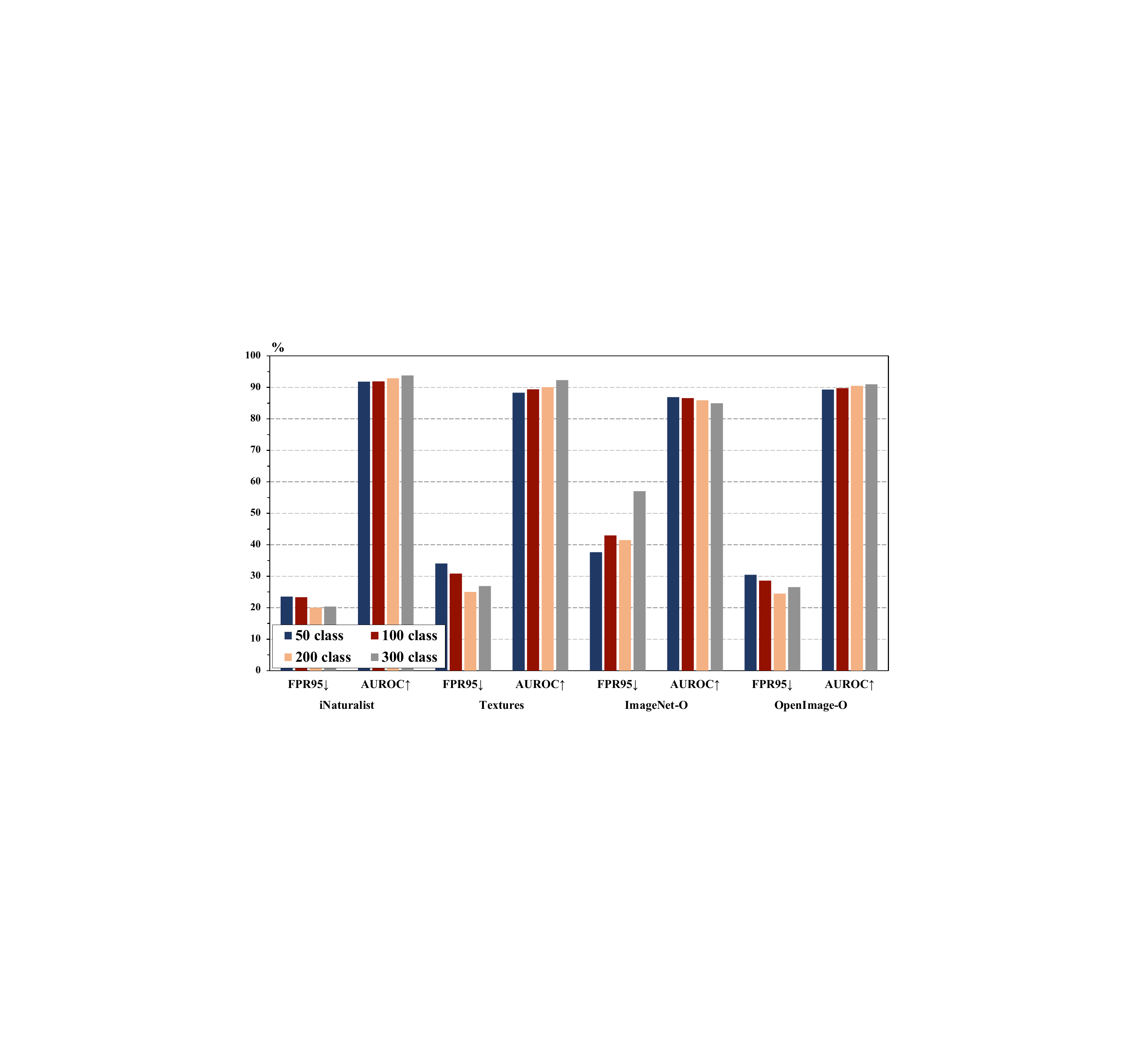}
    }
    \caption{The performance of HVCM is evaluated as the number of InD classes increases across four OOD datasets.}
    \label{fig:classabl}
    % \vspace{-1em}
\end{figure}

{\setlength{\parindent}{0cm}
\textbf{Increasing numbers of InD Categories in ImageNet.}
We test how the OOD detection performance varies with the increasing of object categories in the in-distribution dataset. Following Wang et al.~\cite{haoqi2022vim}, we test on four popular benchmarks and set $C$ to $\{50,100,200,300\}$ respectively. As depicted in Figure \ref{fig:classabl}, the performance of HVCM fluctuates on different datasets with the increasing of the InD object categories, which suggests that the number of categories has little impact on our approach. These results validate our assumption that we only need to model in-distribution image categories, and the out-of-distribution samples can be detected easily then. 
}

\begin{figure}
  \centering
  \scalebox{0.9}{
    \includegraphics[width=\linewidth]{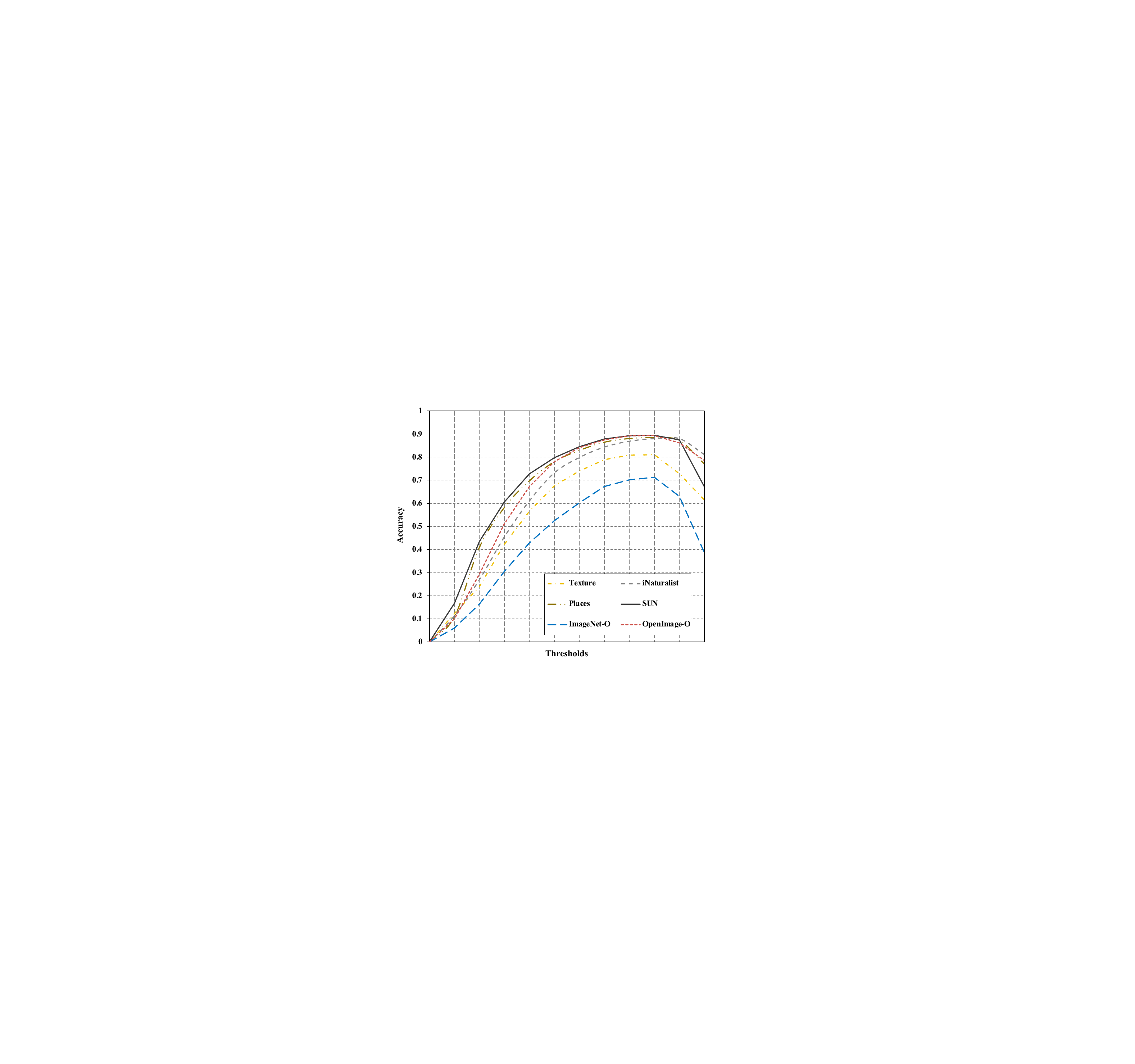}
    }
    \caption{The accuracy of HVCM varies with different thresholds across all OOD datasets.}
    \label{fig:accthres}
    % \vspace{-1em}
\end{figure}

\setlength{\parindent}{0cm}
\textbf{Different thresholds for OOD detection.}
Figure \ref{fig:accthres} illustrates the accuracy of OOD detection for our method across various datasets. On most datasets, our method exhibits the same trend for accuracy variation with thresholds. This indicates that our approach has strong generalization and ideal performance to domains with significant differences. We explain the performance of Imagenet-O as its task difficulty with adversarial samples.

\section{Conclusion}
In this paper, we introduce a hierarchical visual category modeling scheme for out-of-distribution detection, which combines visual representation learning and parameter optimization of probability models. It provides a novel perspective for OOD detection by conducting representation learning and density estimation end-to-end. By modeling visual categories with mixtures of Gaussian models, we describe visual categories in very complex distribution and don't rely on outlier training data to perform OOD detection. Experiments demonstrate that the proposed method outperforms state-of-the-art algorithms clearly and does not hinder the discriminative ability of deep features.

\noindent\textbf{Limitations.} However, our method needs to map deep features into high-dimensional attribute spaces and build plentiful Gaussian mixture models. These Gaussian mixture models bring a lot of computational costs and make the inference of the OOD detector become inefficient. Thus, simplifying the probability models and accelerating the inference process will be the future direction.

\section{Acknowledgment}
This work was supported by National Natural Science Foundation of China (No.62072112), Scientific and Technological innovation action plan of Shanghai Science and Technology Committee (No.22511102202), National Key R\&D Program of China (2020AAA0108301), National Natural Science Foundation of China under Grant (No. 62106051), Shanghai Pujiang Program (No. 21PJ1400600), and National Key R\&D Program of China (2022YFC3601405).

{\small
\bibliographystyle{ieee_fullname}
\bibliography{hvcmood}
}

\end{document}